\documentclass[11pt,a4paper]{article}
\usepackage[hyperref]{acl2020}
\usepackage{times}
\usepackage{latexsym}
\usepackage{graphicx}
\usepackage{textcomp}

\usepackage{comment}
\usepackage{microtype}

\aclfinalcopy 

\title{MLSUM: The Multilingual Summarization Corpus}
\author{Thomas Scialom$^{\star \ddagger}$,  Paul-Alexis Dray$^{\star}$, Sylvain Lamprier$^{\ddagger}$, Benjamin Piwowarski$^{\diamond \ddagger}$, Jacopo Staiano$^{\star}$ \\
$^\diamond$ CNRS, France\\
$^\ddagger$ Sorbonne Universit\'e, CNRS, LIP6, F-75005 Paris, France\\
$^\star$ reciTAL, Paris, France \\
  {\tt \{thomas,jacopo,paul-alexis\}@recital.ai} \\
  {\tt \{sylvain.lamprier,benjamin.piwowarski\}@lip6.fr} \\}

\date{}

\begin{document}
\maketitle
\begin{abstract}
We present MLSUM, the first large-scale MultiLingual SUMmarization dataset. Obtained from online newspapers, it contains 1.5M+ article/summary pairs in five different languages -- namely, French, German, Spanish, Russian, Turkish. Together with English newspapers from the popular CNN/Daily mail dataset, the collected data form a large scale multilingual dataset which can enable new research directions for the text summarization community.
We report cross-lingual comparative analyses based on  state-of-the-art systems. These highlight existing biases which motivate the use of a multi-lingual dataset.

\end{abstract}

\section{Introduction}
The document summarization task requires several complex language abilities: understanding a long document, discriminating what is relevant, and writing a short synthesis. Over the last few years, advances in deep learning applied to NLP have contributed to the rising popularity of this task among the research community \cite{see2017get, kryscinski2018improving, scialom-etal-2019-answers}.
As with other NLP tasks, the great majority of available datasets for summarization  are in English, and thus most research efforts focus on the English language. The lack of multilingual data is partially countered by the application of transfer learning techniques enabled by the availability of pre-trained multilingual language models. This approach has recently established itself as the \emph{de-facto} paradigm in NLP \cite{guzman2019flores}.

Under this paradigm, for encoder/decoder tasks, a language model can first be pre-trained on a large corpus of texts in multiple languages. Then, the model is fine-tuned in one or more \emph{pivot} languages for which the task-specific data are available. At inference, it can still be applied to the different languages seen during the pre-training. 
Because of the dominance of English for large scale corpora, English naturally established itself as a pivot for other languages. The availability of multilingual pre-trained models, such as BERT multilingual (M-BERT), allows to build 
models for target languages different from training data. 
However, previous works reported a significant performance gap between English and the target language, e.g. for classification \cite{conneau2018xnli} and Question Answering \cite{lewis2019mlqa} tasks. A similar approach has been recently proposed for summarization \cite{chi2019cross} obtaining, again, a lower performance than for English.

For specific NLP tasks, recent research efforts have produced evaluation datasets in several target languages, allowing to evaluate the progress of the field in zero-shot scenarios. Nonetheless, those approaches are still bound to using training data in a \emph{pivot} language for which a large amount of annotated data is available, usually English.
This prevents investigating, for instance, whether a given model is as fitted for a specific language as for any other.  Answers to such research questions represent valuable information to improve model performance for low-resource languages.

In this work, we aim to fill this gap for the automatic summarization task by proposing a large-scale MultiLingual SUMmarization (MLSUM) dataset. The dataset is built from online news outlets, and contains over 1.5M article-summary pairs in 5 languages: French, German, Spanish, Russian, and Turkish, which complement an already established  summarization dataset in English. 

The contributions of this paper can be summarized as follows:
\begin{enumerate}
    \item We release the first large-scale multilingual summarization dataset;
    \item We provide strong baselines from multilingual abstractive text generation models;
    \item We report a comparative cross-lingual analysis of the results obtained by different approaches. 
\end{enumerate}

\section{Related Work}

\subsection{Multilingual Text Summarization}

Over the last two decades, several research works have focused on multilingual text summarization. \citet{radev2002evaluation} developed MEAD, a multi-document summarizer that works for both English and Chinese. \citet{litvak2010new} proposed to improve multilingual summarization using a genetic algorithm. A community-driven initiative, MultiLing \cite{giannakopoulos2015multiling}, benchmarked summarization systems on multilingual data. While the MultiLing benchmark covers 40 languages, it provides relatively few examples (10k in the 2019 release). Most proposed approaches, so far, have used an extractive approach given the lack of a multilingual corpus to train abstractive models \cite{duan2019zero}.

More recently, with the rapid progress in automatic translation and text generation, abstractive methods for multilingual summarization have been developed. \citet{ouyang2019robust} proposed to learn summarization models for three low-resource languages (Somali, Swahili, and Tagalog), by using an automated translation of the New York Times dataset..  
Although this showed only slight improvements over a baseline which considers translated outputs of an English summarizer, results remain still far from human performance. Summarization models from translated data usually under-perform, as translation biases add to the difficulty of summarization. 

Following the recent trend of using multi-lingual pre-trained models for NLP tasks, such as Multilingual
BERT (M-BERT) \cite{pires2019multilingual}\footnote{\url{https://github.com/google-research/bert/blob/master/multilingual.md}} or XLM \cite{lample2019cross}, \citet{chi2019cross} proposed to fine-tune the models for summarization on English training data. The assumption is that the summarization skills learned from English data can transfer to other languages on which the model has been pre-trained. However a significant performance gap between English and the target language is observed following this process. This emphasizes the crucial need of multilingual training data for summarization.  

\subsection{Existing Multilingual Datasets}
\label{Existing_Multilingual_datasets}

The research community has produced several multilingual datasets for tasks other than summarization. We report two recent efforts below, noting that both \emph{i)} rely on human translations, and \emph{ii)} only provide evaluation data.

\paragraph{The Cross-Lingual NLI Corpus} 
The SNLI corpus \cite{bowman2015large} is a large scale dataset for natural language inference (NLI). It is composed of a collection of 570k human-written English sentence pairs, associated with their label, entailment, contradiction, or neutral. The Multi-Genre Natural Language Inference (MultiNLI) corpus is an extension of SNLI, comparable in size, but including a more diverse range of text. 
\citet{conneau2018xnli} introduced the Cross-Lingual NLI Corpus (XNLI) to evaluate transfer learning from English to other languages: based on MultiNLI, a collection of 5,000 test and 2,500 dev pairs were translated by humans in 15 languages.

\paragraph{MLQA}
Given a paragraph and a question, the Question Answering (QA) task consists in providing the correct answer. Large scale datasets such as \cite{rajpurkar2016squad,choi2018quac,trischler2016newsqa} have driven fast progress.\footnote{For instance, see the SQuAD leaderboard:  \url{rajpurkar.github.io/SQuAD-explorer/}} However, these datasets are only in English. To assess how well models perform on other languages, \citet{lewis2019mlqa} recently proposed MLQA, an evaluation dataset for cross-lingual extractive QA composed of 5K QA instances in 7 languages.

\paragraph{XTREME}
The Cross-lingual TRansfer Evaluation of Multilingual Encoders benchmark covers 40 languages over 9 tasks. The summarization task is not included in the benchmark.

\paragraph{XGLUE}
In order to train and evaluate their performance across a diverse set of cross-lingual tasks, \citet{liang2020xglue}  recently released XGLUE, covering both Natural Language Understanding and Generation scenarios. While no summarization task is included, it comprises a News Title Generation task: the data is crawled from a commercial news website and provided in form of article-title pairs for 5 languages (German, English, French, Spanish and Russian).

\subsection{Existing Summarization datasets}
\label{Existing_Summarization_datasets}

We describe here the main available corpora for text summarization.  

\paragraph{Document Understanding Conference} 
Several small and high-quality summarization datasets in English  \cite{harman2004effects, dang2006duc} have been produced in the context of the Document Understanding Conference (DUC).\footnote{\url{http://duc.nist.gov/}}
They are built by associating newswire articles with corresponding human summaries. A distinctive feature of the DUC datasets is the availability of multiple reference summaries: this is a valuable characteristic since, as found by \citet{rankel-etal-2013-decade}, the correlation between qualitative and automatic metrics, such as ROUGE \cite{lin2004rouge}, decreases significantly when only a single reference is given. However, due to the small number of training data available, DUC datasets are often used in a domain adaptation setup for models first trained on larger datasets such as Gigaword, CNN/DM \cite{nallapati2016abstractive, see2017get} or with unsupervised methods \cite{dorr2003hedge, mihalcea2004textrank, barrios2016variations}.

\paragraph{Gigaword} 
Again using newswire as source data, the english Gigaword \cite{napoles2012annotated, rush2015neural, chopra2016abstractive} corpus is characterized by its large size and the high diversity in terms of sources. Since the samples are not associated with human summaries, prior works on summarization have trained models to generate the headlines of an article, given its incipit, which induces various  biases for learning models.

\paragraph{New York Times Corpus}  
This large corpus for summarization consists of hundreds of thousand of articles from The New York Times\cite{sandhaus2008new}, spanning over 20 years. The articles are paired with summaries written by library scientists. 
Although \cite{grusky2018newsroom} found indications of bias towards extractive approaches, several research efforts have used this dataset for summarization \cite{hong2014improving,durrett2016learning,paulus2017deep}.

\paragraph{CNN / Daily Mail}
One of the most commonly used dataset for summarization \cite{nallapati2016abstractive, see2017get, paulus2017deep, dong2019unified}, although originally built for Question Answering tasks \cite{hermann2015teaching}. It consists of English articles from the CNN and The Daily Mail associated with bullet point highlights from the article. 
When used for summarization, the bullet points are typically concatenated into a single summary. 

\paragraph{NEWSROOM}
Composed of 1.3M articles \cite{grusky2018newsroom}, and featuring high diversity in terms of publishers, the summaries associated with English news articles were extracted from the Web pages metadata: they were originally written to be used in search engines and social media.

\paragraph{BigPatent}
\citet{sharma2019bigpatent} collected 1.3 million U.S. patent documents, across several technological areas, using the Google Patents Public Datasets. The patents abstracts are used as target summaries.

\paragraph{LCSTS}
The Large Scale Chinese Short Text Summarization Dataset \cite{hu2015lcsts} is built from 2 million short texts from the Sina Weibo microblogging platform. They are paired with summaries given by the author of each text. The dataset includes 10k summaries which were manually scored by human for their relevance.

\section{MLSUM}
As described above, the vast majority of summarization datasets are in English. For Arabic, there exist the Essex Arabic Summaries Corpus (EASC) \cite{el2010using} and KALIMAT \cite{el2013kalimat}; those comprise circa 1k and 20k samples, respectively. \citet{pontes2018new} proposed a corpus of few hundred samples for Spanish, Portuguese and French summaries. To our knowledge, the only large-scale non-English summarization dataset is the Chinese LCSTS \cite{hu2015lcsts}. 
With the increasing interest for cross-lingual models, the NLP community have recently released multilingual evaluation datasets, targeting classification (XNLI) and  QA \cite{lewis2019mlqa} tasks, as described in \ref{Existing_Multilingual_datasets}, though still no large-scale dataset is avaulable for document summarization.

To fill this gap we introduce MLSUM, the first large scale multilingual summarization corpus. 
Our corpus provides more than 1.5 millions articles in French (FR), German (DE), Spanish (ES), Turkish (TR), and Russian (RU). Being similarly built from news articles, and  providing a similar amount of training samples per language (except for Russian), as the previously mentioned CNN/Daily Mail, it can effectively serve as a multilingual extension of the CNN/Daily Mail dataset. 

In the following, we first describe the methodology used to build the corpus. We then report the corpus statistics and finally interpret the performances of baselines and state-of-the-art models.

\subsection{Collecting the Corpus} 
The CNN/Daily Mail (CNN/DM) dataset (see Section \ref{Existing_Summarization_datasets}) is arguably the most used large-scale dataset for summarization. Following the same methodology, we consider news articles as the text input, and their paired \emph{highlights/description} as the summary. For each language, we selected an online newspaper which met the following requirements: 
\begin{enumerate}
\item Being a \emph{generalist} newspaper: ensuring that a broad range of topics is represented for each language allows to minimize the risk of training topic-specific models, a fact which would hinder comparative cross-lingual analyses of the models.
\item Having a large number of articles in their public online archive.
\item Providing human written highlights/summaries for the articles that can be extracted from the HTML code of the web page.
\end{enumerate}

After a careful preliminary exploration, we selected the online version of the following newspapers: 
\begin{itemize}
    \item Le Monde\footnote{\url{www.lemonde.fr}} (French)
    \item S{\"u}ddeutsche Zeitung\footnote{\url{www.sueddeutsche.de}} (German)
    \item El Pais\footnote{\url{www.elpais.com}} (Spanish)
    \item Moskovskij Komsomolets\footnote{\url{www.mk.ru}} (Russian)
    \item Internet Haber\footnote{\url{www.internethaber.com}} (Turkish)
\end{itemize}

For each outlet, we crawled archived articles from 2010 to 2019. We applied one simple filter: all the articles shorter than 50 words or summaries shorter than 10 words are discarded, so as to avoid articles containing mostly audiovisual content. Each article was archived on the Wayback Machine,\footnote{\url{web.archive.org}, using \url{https://github.com/agude/wayback-machine-archiver}} allowing interested research to re-build or extend MLSUM. 
We distribute the dataset as a list of immutable snapshot URLs of the articles, along with the accompanying corpus-construction code,\footnote{\url{https://github.com/recitalAI/MLSUM}} allowing to replicate the parsing and preprocessing procedures we employed.
This is due to legal reasons: the content of the articles is copyrighted and redistribution might be seen as infringing of publishing rights. Nonetheless, we make available, upon request, an exact copy of the dataset used in this work. A similar approach has been adopted for several dataset releases in the recent past, such as Question Answering Corpus \cite{nips15_hermann} or XSUM \cite{xsum-emnlp}.

Further, we provide recommended train/validation/test splits following a chronological ordering based on the articles' publication dates. In our experiments below, we train/evaluate the models on the training/test splits obtained in this manner. Specifically, we use: data from 2010 to 2018, included, for training; data for 2019 (\texttildelow10\% of the dataset) for validation (up to May 2019) and test (May-December 2019). While this choice is arguably more challenging, due to the possible emergence of new topics over time, we consider it as the realistic scenario a successful summarization system should be able to deal with. Incidentally, this also bring the advantage of excluding most cases of leakage across languages: it prevents a model, for instance, from seeing a training sample describing an important event in one language, and then being submitted for inference a similar article in another language, published around the same time and dealing with the same event.

\subsection{Dataset Statistics}

\begin{table*}[!ht]
    \centering
    \begin{tabular}{l|l|l|l|l|l||r}
         &  FR & DE	& ES & RU & TR & EN\\
         \hline
Dataset size & 424,763 & 242,982 & 290,645 & 27,063 & 273,617 & 311,971\\
Training set size & 392,876 & 220,887 & 266,367 & 25,556 & 249,277 & 287,096 \\
\hline
Mean article length & 632.39 & 570.6 & 800.50 & 959.4 & 309.18 & 790.24\\
Mean summary length & 29.5 & 30.36 & 20.71 & 14.57 & 22.88 & 55.56\\
Compression Ratio    & 21.4 & 18.8 & 38.7 & 65.8 & 13.5 & 14.2 \\
Novelty (1-gram) &15.21&14.96&15.34&30.74&28.90&9.45\\
\hline
Total Vocabulary Size & 1,245,987 & 1,721,322 & 1,257,920 & 649,304 & 1,419,228 & 875,572\\
Occurring 10+ times & 233,253 & 240,202 & 229,033 & 115,144 & 248,714 & 184,095\\

    \end{tabular}
    \caption{Statistics for the different languages. \emph{EN} refers to CNN/Daily Mail and is reported for comparison purposes. Article and summary lengths are computed in words. Compression ratio is computed as the ratio between article and summary length. Novelty is the percentage of words in the summary that were not in the paired article. Total Vocabulary is the total number of different words and Occurring 10+, the total number of words occurring 10+ times.  }
    \label{tab:table_description_dataset}
\end{table*}
We report statistics for each language in MLSUM in Table~\ref{tab:table_description_dataset}, including those computed on the CNN/Daily Mail dataset (English) for quick comparison. MLSUM provides a comparable amount of data for all languages, with the exception of Russian with ten times less training samples. Important characteristics for summarization datasets are the length of articles and summaries, the vocabulary size, and a proxy for abstractiveness, namely the percentage of novel n-grams between the article and its human summary. From Table~\ref{tab:table_description_dataset}, we observe that Russian summaries are the shortest as well as the most abstractive. 

Coupled with the significantly lower amount of articles available from its online source, the task can be seen as more challenging for Russian than for the other languages in MLSUM. Conversely, similar characteristics are shared among other languages, for instance French and German.

\subsection{Topic Shift}
\begin{figure*}[!ht]
    \centering
    \includegraphics[width=.49\linewidth]{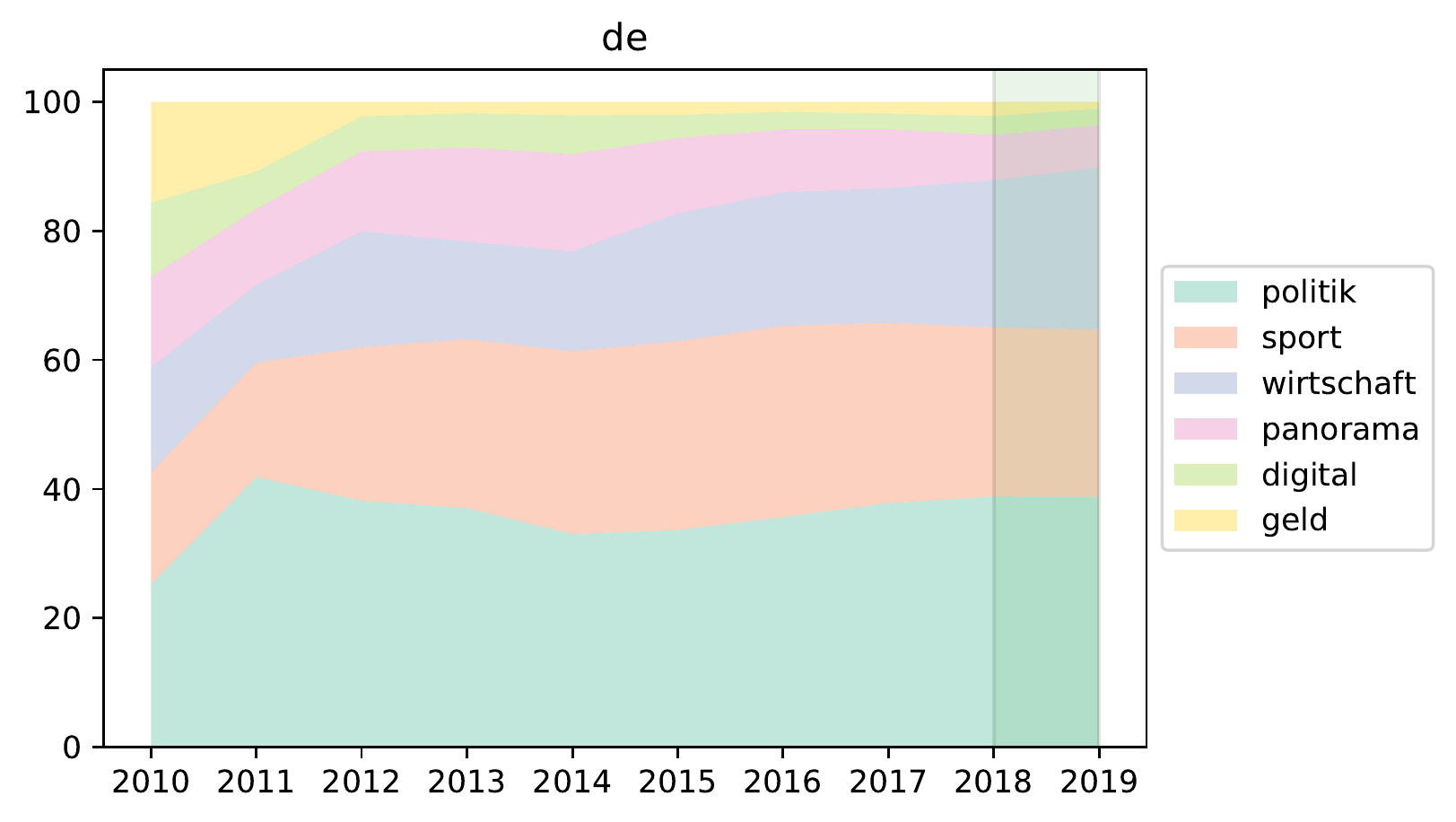}
    \includegraphics[width=.49\linewidth]{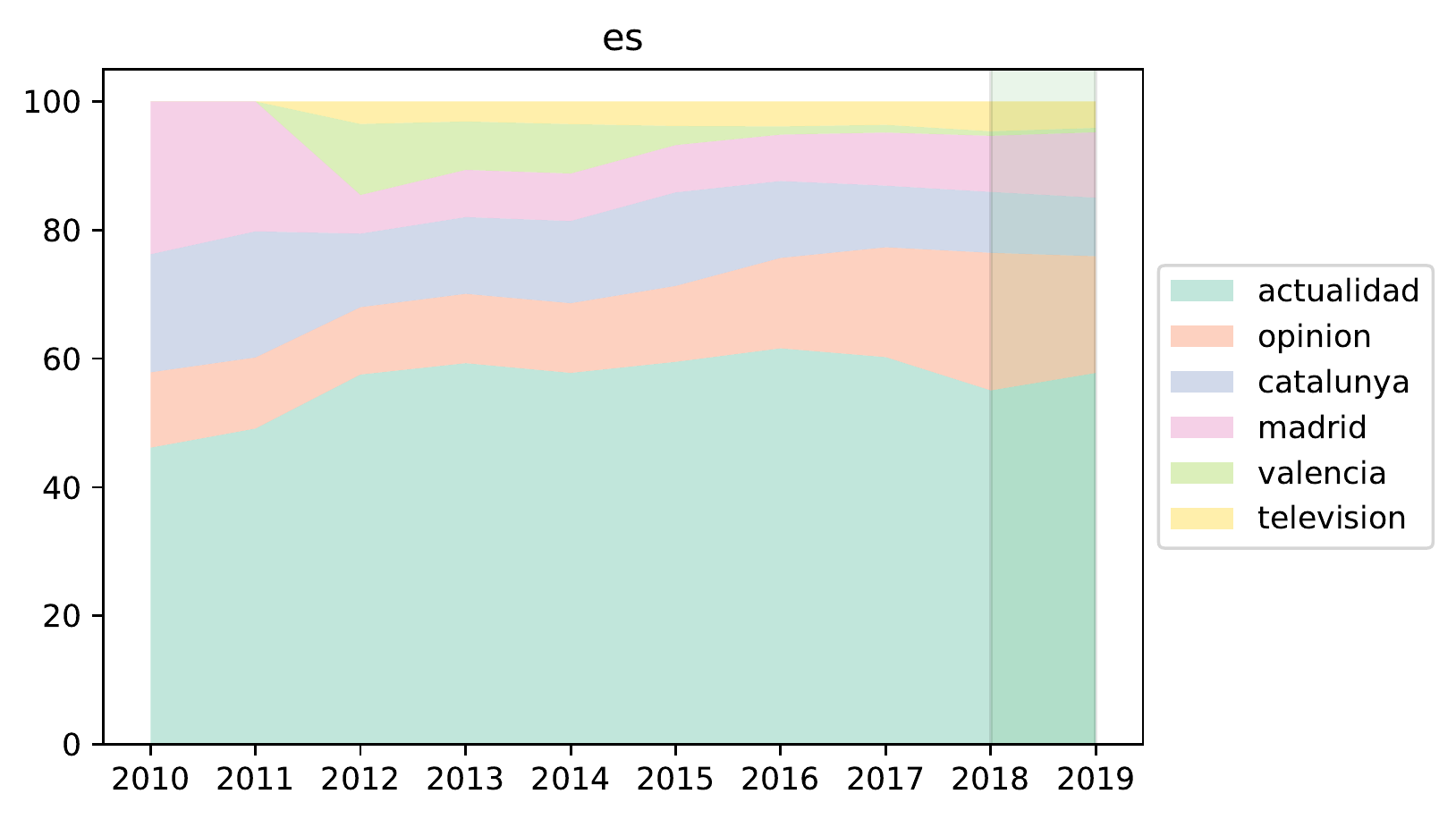}
    \includegraphics[width=.49\linewidth]{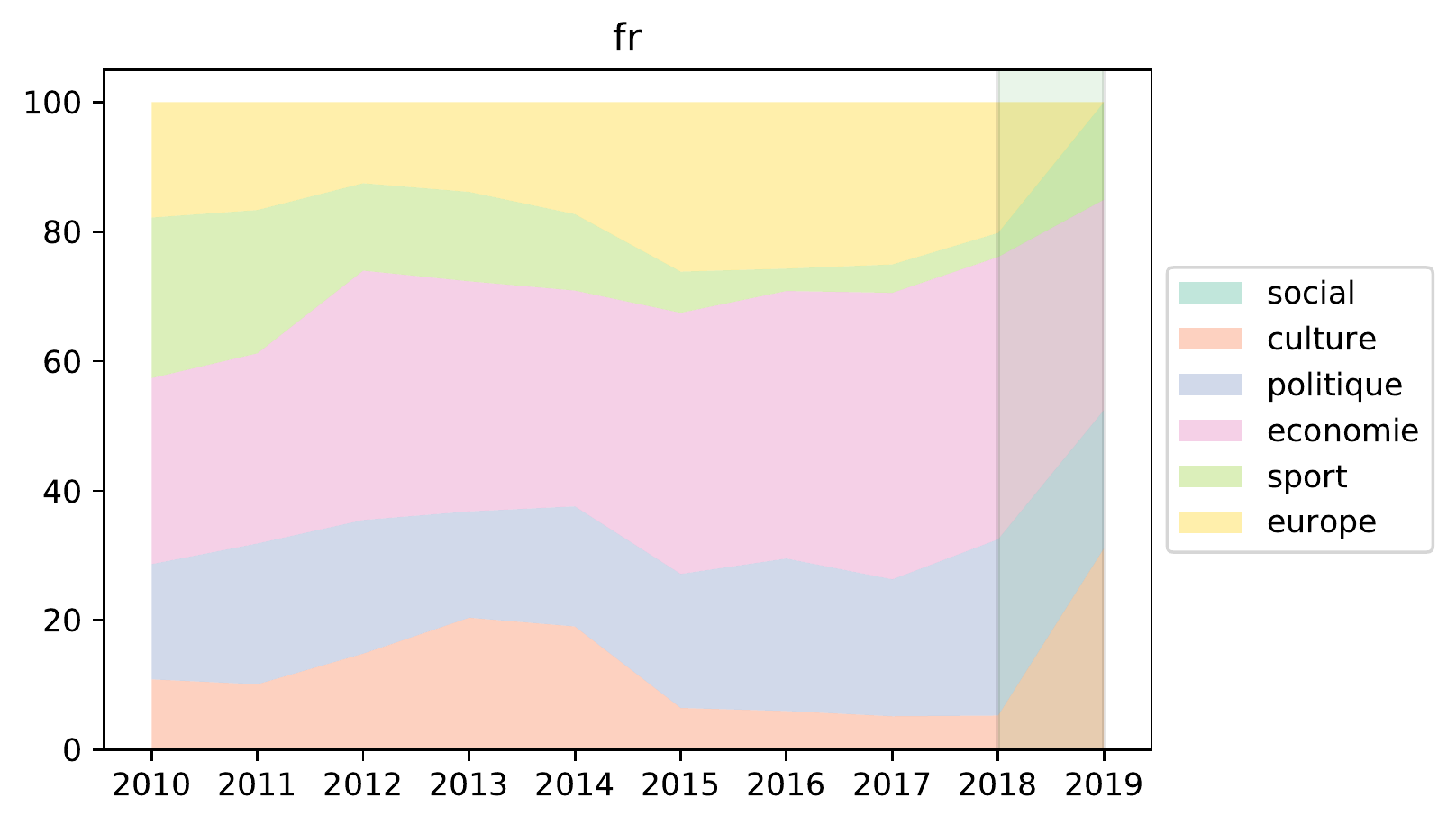}
    \includegraphics[width=.49\linewidth]{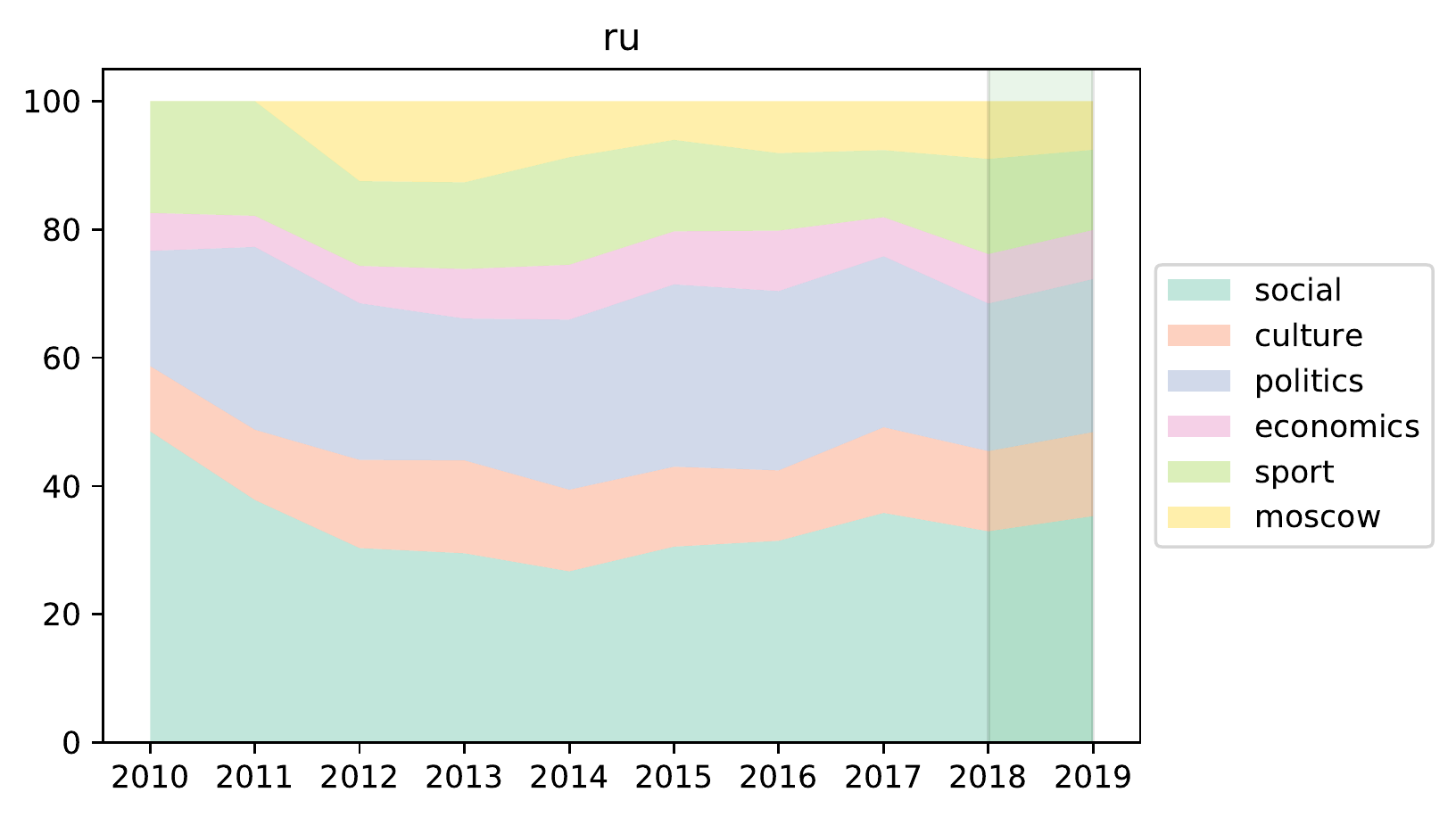}
    \caption{Distribution of topics for German (top-left), Spanish (top-right), French (bottom-left) and Russian (bottom-right), grouped per year. The shaded area for 2019 highlights validation and test data.}
    \label{fig:topic_stackplot}
\end{figure*}

With the exception of Turkish, the article URLs in MLSUM allow to identify a category for a given article. In Figure~\ref{fig:topic_stackplot} we show the shift over categories among time. In particular, we plot the 6 most frequent categories per language.

\section{Models}
We experimented on MLSUM with the established models and baselines described below.
Those include supervised and unsupervised methods, extractive and abstractive models. For all the experiments, we train models on a per-language basis. We used the recommended hyperparameters for all languages, in order to facilitate assessing the robustness of the models. We also tried to train one model with all the languages mixed together, but we did not see any significant difference of performance. 

\subsection{Extractive summarization models}
\label{sec:models}

\paragraph{Oracle} 
Extracts the sentences, within the input text, that maximise a given metric (in our experiments, ROUGE-L) given the reference summary.
It is an indication of the maximum one could achieve with extractive summarization. In this work, we rely on the implementation of \citet{narayan-etal-2018-ranking}. 

\paragraph{Random} 
In order to elaborate and compare the performances of the different models across languages, it is useful to include an unbiased model as a point of reference. To that purpose, we define a simple random extractive model that randomly extracts $N$ words from the source document, with $N$ fixed as the average length of the summary. 

\paragraph{Lead-3} 
Simply selects the three first sentences from the input text.  \citet{sharma2019bigpatent}, among others, showed that this is a robust baseline for several summarization datasets such as CNN/DM, NYT and BIGPATENT.

\paragraph{TextRank}
An unsupervised algorithm proposed by \citet{mihalcea2004textrank}. It consists in computing the co-similarities between all the sentences in the input text. Then, the most central to the document are extracted and considered as the summary. We used the implementation provided by \citet{DBLP:journals/corr/BarriosLAW16}.

\subsection{Abstractive summarization models}
Most of the models for abstractive summarization are neural sequence to sequence models \cite{sutskever2014sequence}, composed of an encoder that encodes the input text and a decoder that generates the summary.

\paragraph{Pointer-Generator}
\label{par:Pointer-Generator}
\citet{see2017get} proposed the addition of the copy mechanism \cite{vinyals2015pointer} on top of a sequence to sequence LSTM model. This mechanism allows to efficiently copy out-of-vocabulary tokens, leveraging attention \cite{bahdanau2014neural} over the input. We used the publicly available OpenNMT implementation\footnote{\url{opennmt.net/OpenNMT-py/Summarization.html}} with the default hyper-parameters. However, to avoid biases, we limited the preprocessing as much as possible and did not use any sentence separators, as recommended for CNN/DM. This explains the relatively lower reported ROUGE, compared to the model with the full preprocessing. 
 
\paragraph{M-BERT} 
Encoder-decoder Transformer  architectures are a very popular choice for text generation.
Recent research efforts have adapted large pretrained self-attention based models for text generation \cite{peters2018deep,radford2018improving,devlin2019bert}.

In particular, \citet{liu2019text} added a randomly initialized decoder on top of BERT. Avoiding the use of a decoder, \citet{dong2019unified} proposed to instead add a decoder-like mask during the pre-training to \emph{unify} the language models for both encoding and generating. Both these approaches achieved SOTA results for summarization. 
In this paper, we only report results obtained following \citet{dong2019unified}, as in preliminary experiments we observed that a simple multilingual BERT (M-BERT), with no modification, obtained comparable performance on the summarization task.

\begin{table*}[!htb]
    \centering
    \begin{tabular}{l|r|r|r|r|r||r}
         &FR&DE&ES&RU&TR&EN\\
         \hline
Oracle&37.69&52.3&35.78&29.80&45.78&53.6\\
Random&11.88&10.22&12.63&6.7&11.29&11.23\\
TextRank&12.61&13.26&9.5&3.28&21.5&28.61\\
Lead\_3&19.69&33.09&13.7&5.94&28.9&35.2\\
Pointer-Generator&23.58&35.08&17.67&5.71&32.59&33.32\\
M-BERT&25.09&42.01&20.44&9.48&32.94&35.41\\
    \end{tabular}
    
    \begin{tabular}{l|r|r|r|r|r||r}
         &FR&DE&ES&RU&TR&EN\\
         \hline
Oracle&24.73&31.67&26.45&20.32&26.42&29.99\\
Random&7.54&6.67&6.48&2.5&6.29&10.56\\
TextRank&10.77&13.01&11.14&3.79&14.36&20.37\\
Lead\_3&12.62&23.85&10.26&5.77&20.24&21.16\\
Pointer-Generator&14.07&24.41&13.17&5.69&19.78&20.78\\
M-BERT&15.07&26.47&14.92&6.77&26.26&22.16\\
    \end{tabular}
    \caption{ROUGE-L (top) and METEOR (bottom) results obtainedby the models described in \ref{sec:models} on the different proposed datasets .}
    \label{tab:results}
\end{table*}

\section{Evaluation Metrics}
\label{sec:evaluation_metrics}
\paragraph{ROUGE}
Arguably the most often reported set of metrics in summarization tasks, the  Recall-Oriented Understudy for Gisting Evaluation \cite{lin2004rouge} computes the number of n-grams similar between the evaluated summary and the human reference summary. 

\paragraph{METEOR}
The Metric for Evaluation of Translation with Explicit ORdering \cite{banerjee2005meteor} was designed for the evaluation of machine translation output. It is based on the harmonic mean of unigram precision and recall, with recall weighted higher than precision. METEOR is often reported in summarization papers \cite{see2017get,dong2019unified} in addition to ROUGE.

\paragraph{Novelty}
Because of their use of copy mechanisms, some abstractive models have been reported to rely too much on extraction \cite{see2017get,kryscinski2018improving}. Hence, it became a common practice to report the percentage of novel n-grams produced within the generated summaries. 

\paragraph{Neural Metrics}
Several approaches based on neural models have been recently proposed. Recent works \cite{eyal2019question,scialom-etal-2019-answers} have proposed to evaluate summaries with QA based methods: the rationale is that a good summary should answer the most relevant questions about the article. Further, \citet{kryscinski2019evaluating} proposed a discriminator trained to measure the factualness of the summary. While \citet{bohm-etal-2019-better} learned a metric from human annotation. All these models were only trained on English datasets, preventing us to report them in this paper. 
The availability of MLSUM will enable future works to build such metrics in a multilingual fashion.

\begin{figure*}[!htb]
    \centering
    \includegraphics[width=.32\linewidth]{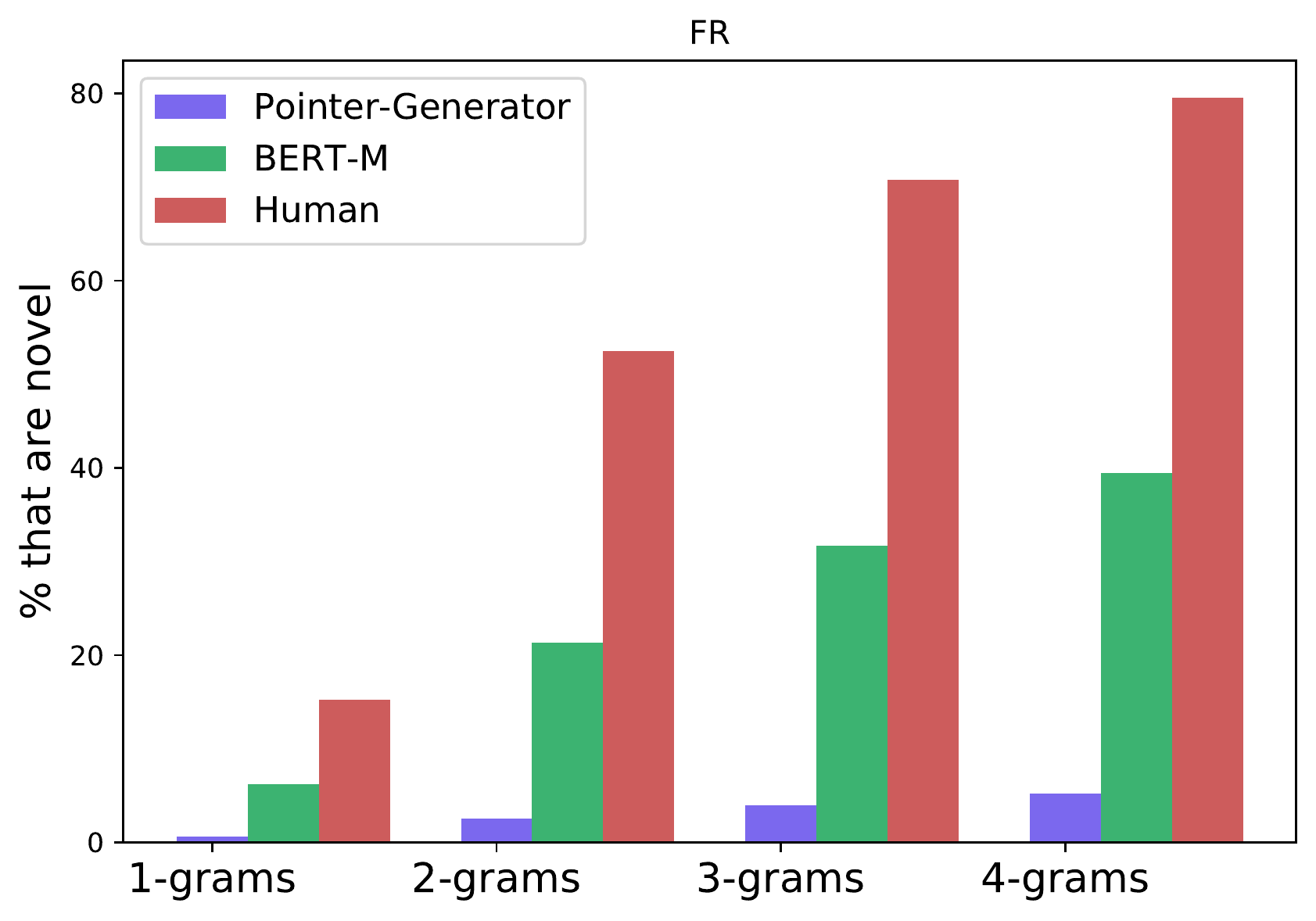}
    \includegraphics[width=.32\linewidth]{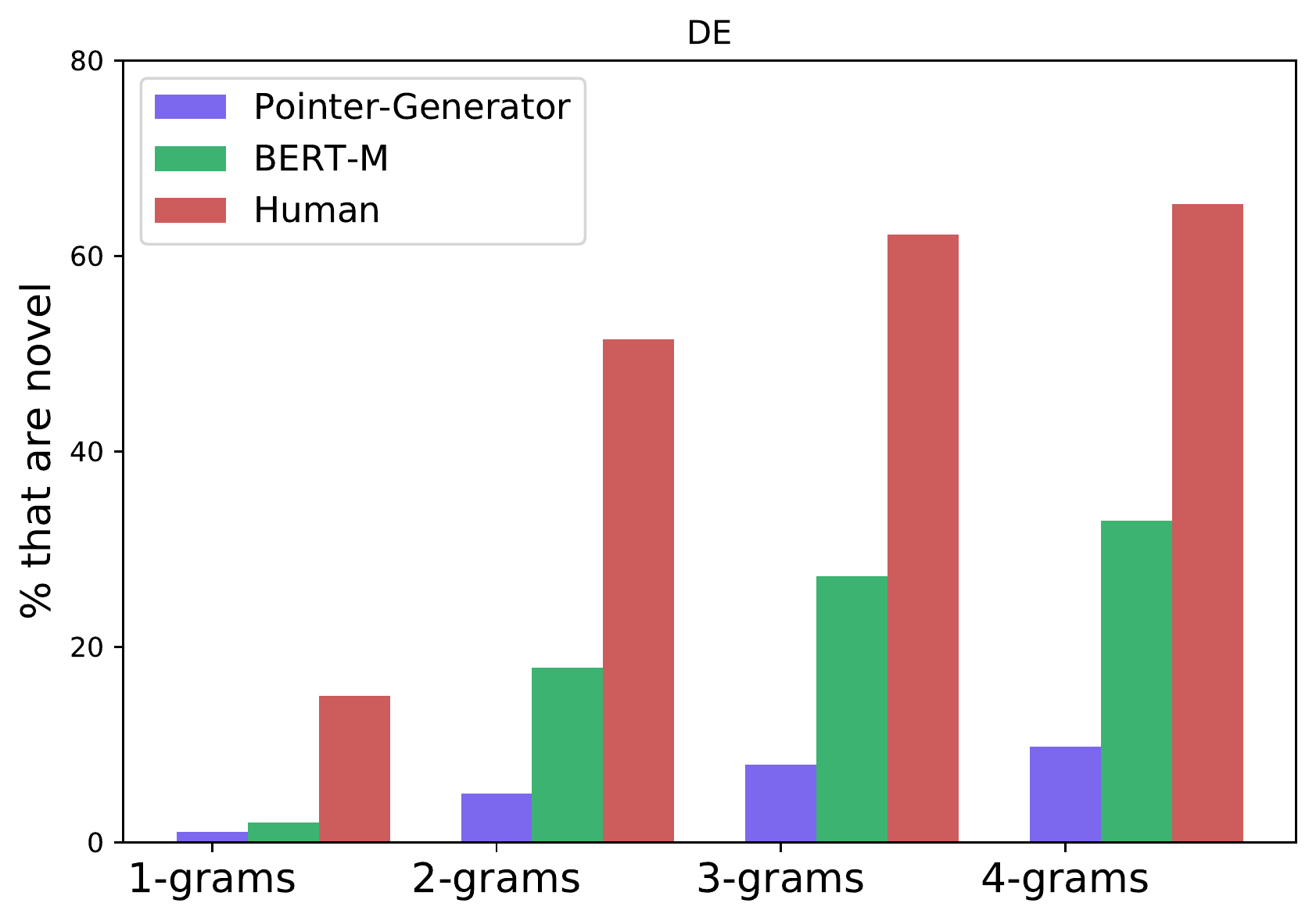}
    \includegraphics[width=.32\linewidth]{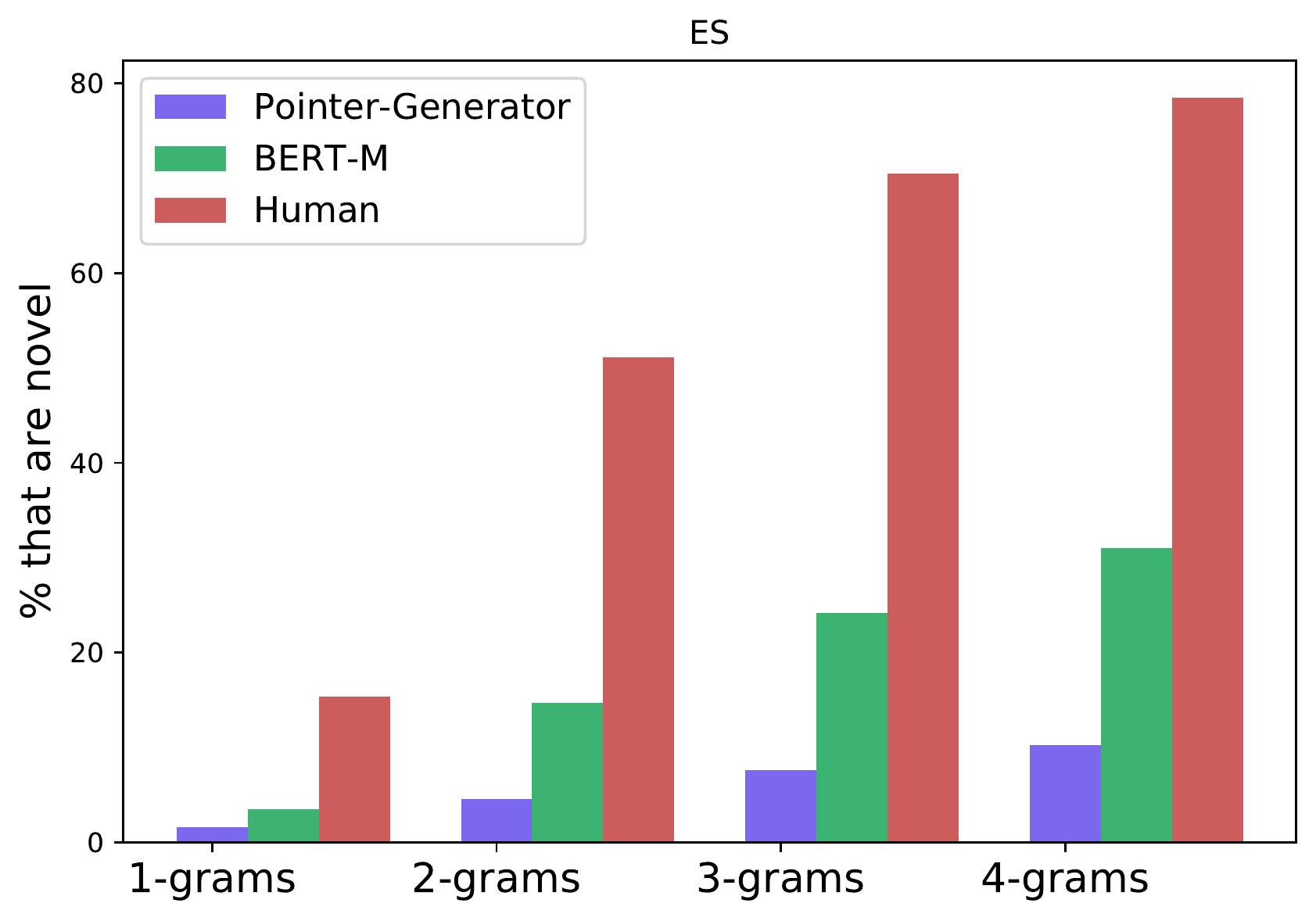}
    \includegraphics[width=.32\linewidth]{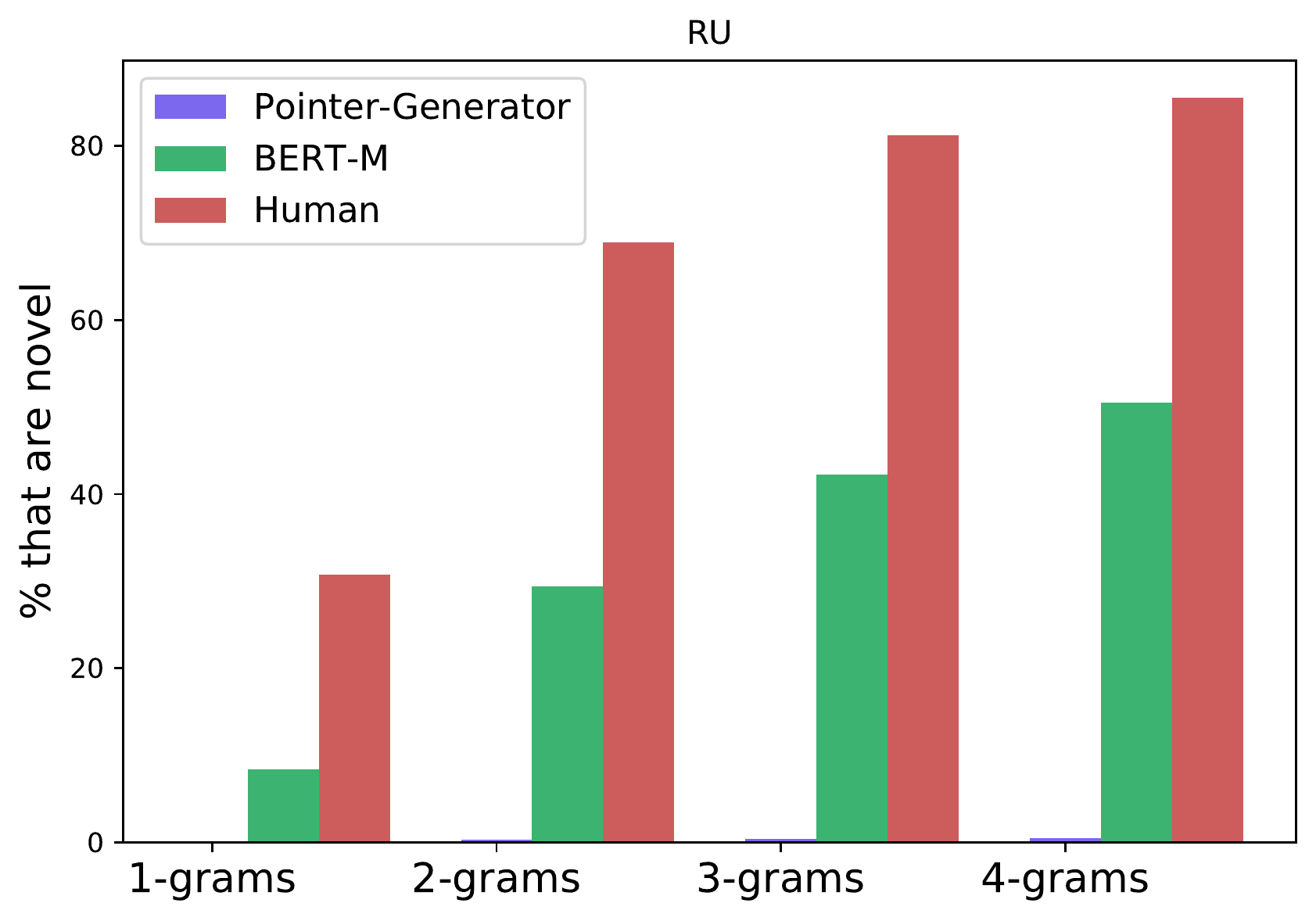}
    \includegraphics[width=.32\linewidth]{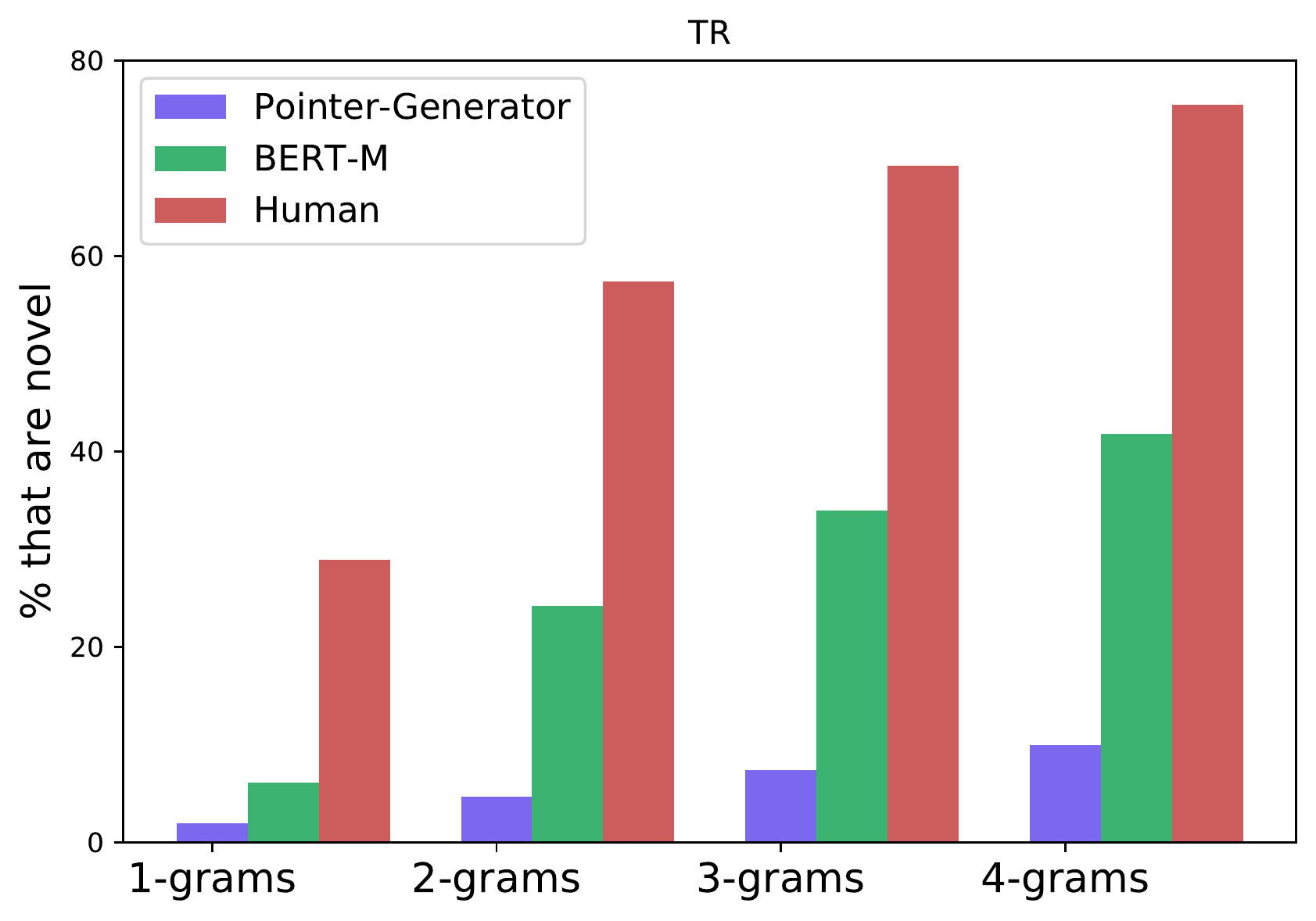}
    \includegraphics[width=.32\linewidth]{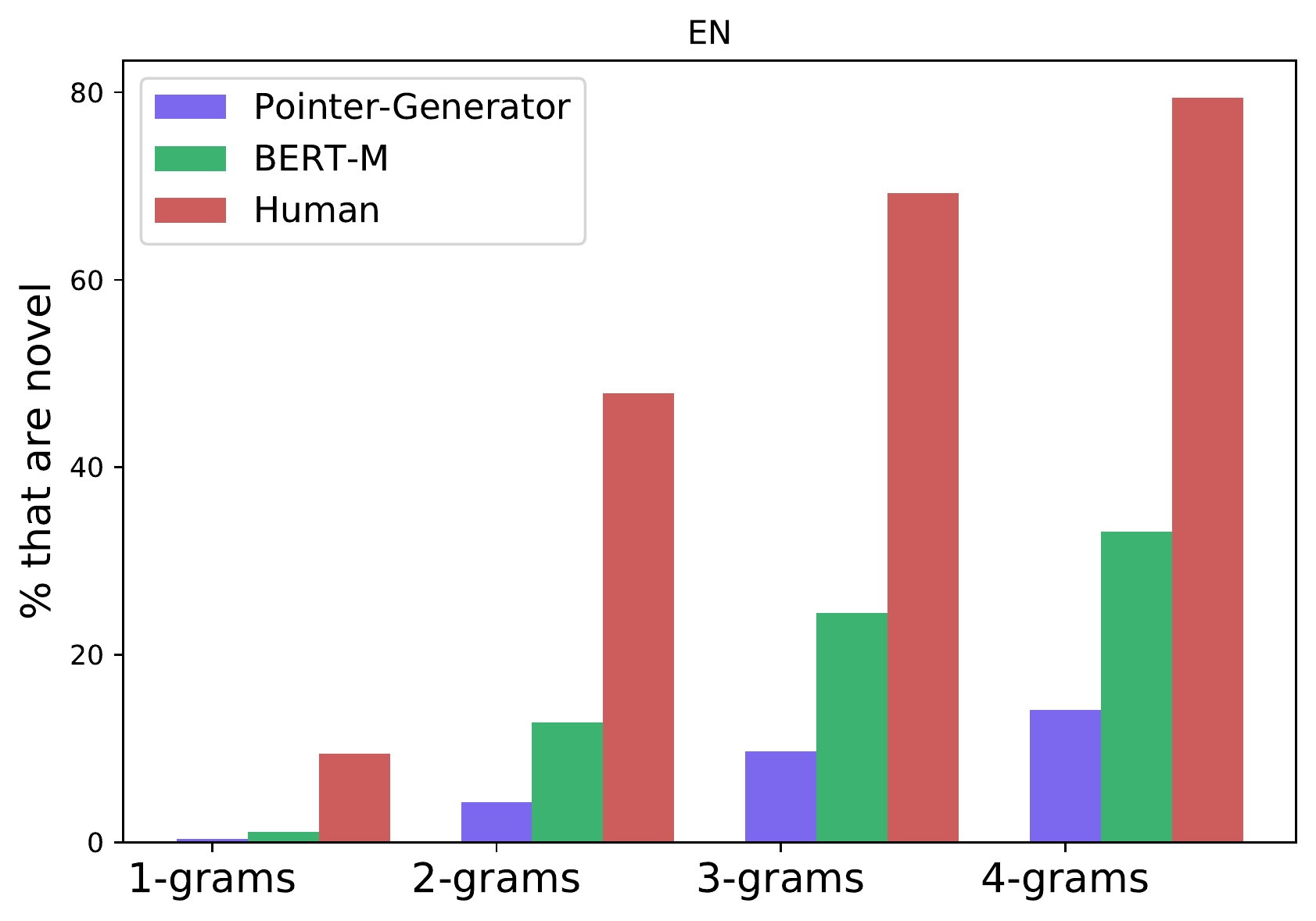}
    \caption{Percentage of novel n-grams for different abstractive models (neural and human), for the 6 datasets.}
    \label{fig:novelty}
\end{figure*}

\section{Results and Discussion}

The results presented below allow us to compare the models across languages, and investigate or hypothesize where their performance variations may come from. We can distinguish the following factors to explain differences in the results:
\begin{enumerate}
    \item Differences in the data, independently from the language, such as the structure of the article, the abstractiveness of the summaries, or the quantity of data;
    \item Differences due to the language itself -- either due to metric biases (e.g. due to a different morphological type) or to biases inherent to the model.
\end{enumerate}

While the first fold of differences have more to do with domain adaptation, the second fold motivates further the development of multilingual datasets, since they are the only mean to study such phenomenon. 

Turning to the observed results, we report in Table~\ref{tab:results} the ROUGE-L and METEOR scores obtained by each model for all languages. 
We note that the overall order of systems (for each language) is preserved when using either metric (modulo some swaps between Lead\_3 and Pointer Generator, but with relatively close scores).

\paragraph{Russian, the low-resource language in MLSUM}

For all experimental setups, the performance on Russian is comparatively low.

This can be explained by at least two factors.
First, the corpus is the most abstractive (see Table~\ref{tab:table_description_dataset}, limiting the performance figures obtained for the extractive models (Random, LEAD-3, and Oracle). 
Second, one order of magnitude less training data is available for Russian than for the other MLSUM languages, a fact which can explain the impressive improvement of performance (+66\% in terms of ROUGE-L, see Table~\ref{tab:results}) between a \emph{not pretrained} model (Pointer Generator) and a \emph{pretrained} model (M-BERT).

\subsection{How abstractive are the models?}
We report the novelty (i.e. the percentage of novel words in the summary) in Figure \ref{fig:novelty}. As previous works reported \cite{see2017get}, pointer-generator networks are poorly abstractive, relying too much on their copy mechanism. 
It is particularly true for Russian: the lack of data probably makes it easier to learn to copy than to cope with natural language generation.
As expected, pretrained language models such as M-BERT are consistently more abstractive, and by a large margin, since they are exposed to other texts during pretraining.

\subsection{Model Biases toward Languages}

\paragraph{Consistency among ROUGE scores}

The Random model obtains comparable ROUGE-L scores across all the languages, except for Russian. This can be explained by the aforementioned Russian corpus characteristics: highest novelty, shortest summaries, and longest input documents (see Table~\ref{tab:table_description_dataset}). 

Thus, in the following, for pair-wise language-based comparisons we focus only on scores obtained, by the different models, on French, German, Spanish, and Turkish -- since we cannot draw meaningful interpretations over Russian as compared to other languages. 

\paragraph{Abstractiveness of the datasets}

The Oracle performance can be considered as the upper limit for an extractive model since it extracts the sentences that provide the best ROUGE-L. We can observe that while being similar for English and German, and to some extent Turkish, the Oracle performance is lower for French or Spanish. 

However, as described in figure \ref{tab:table_description_dataset}, the percentage of novel words are similar for German (14.96), French (15.21) and Spanish (15.34). This may indicate that the relevant information to extract from the article is more spread among sentences for Spanish and French than for German. This is confirmed with the results of Lead-3: German and English have a much higher ROUGE-L -- 35.20 and 33.09 -- than French or Spanish -- 19.69 and 13.70.

\paragraph{The case of TextRank}

\begin{table}[!htb]
    \centering
    \begin{tabular}{l|c|c}
         &T/P &B/P\\
         \hline
FR&0.53&1.06\\
DE&0.37&1.20\\
ES&0.53&1.15\\
RU&0.57&1.65\\
TR&0.65&1.01\\
CNN/DM (EN)&1.10&1.06\\
CNN/DM (EN full preprocessing)&0.85&-\\
DUC (EN)&1.21&-\\
NEWSROOM (EN)&1.10&-\\
    \end{tabular}
    \caption{Ratios of Rouge-L: T/P is the ratio of TextRank to Pointer-Generator and B/P is the ratio of M-BERT to Pointer-Generator. The results for CNN/DM-full preprocessing, DUC and NEWSROOM datasets are those reported in Table 2 of \citet{grusky2018newsroom} (Pointer-C in their paper is our Pointer-Generator).}
    \label{tab:ratios}
\end{table}
The TextRank performance varies widely across the different languages, regardless Oracle. It is particularly surprising to see the low performance on German whereas, for this language, Lead-3 has a comparatively higher performance.
On the other hand, the performance on English is remarkably high: the ROUGE-L is 33\% higher than for Turkish, 126\% higher than for French and 200\% higher than for Spanish. We suspect that the TextRank parameters might actually overfit English.

In Table \ref{tab:ratios}, we report the performance ratio between TextRank and Pointer Generator on our corpus, as well as on CNN/DM and two other English corpora (DUC and NewsRoom). TextRank has a performance close to the Pointer Generator on English corpora (ratio between 0.85 to 1.21) but not in other languages (ratio between 0.37 to 0.65). 

This suggests that this model, despite its generic and unsupervised nature, might be highly biased towards English.

\begin{figure} 
    \includegraphics[width=\linewidth]{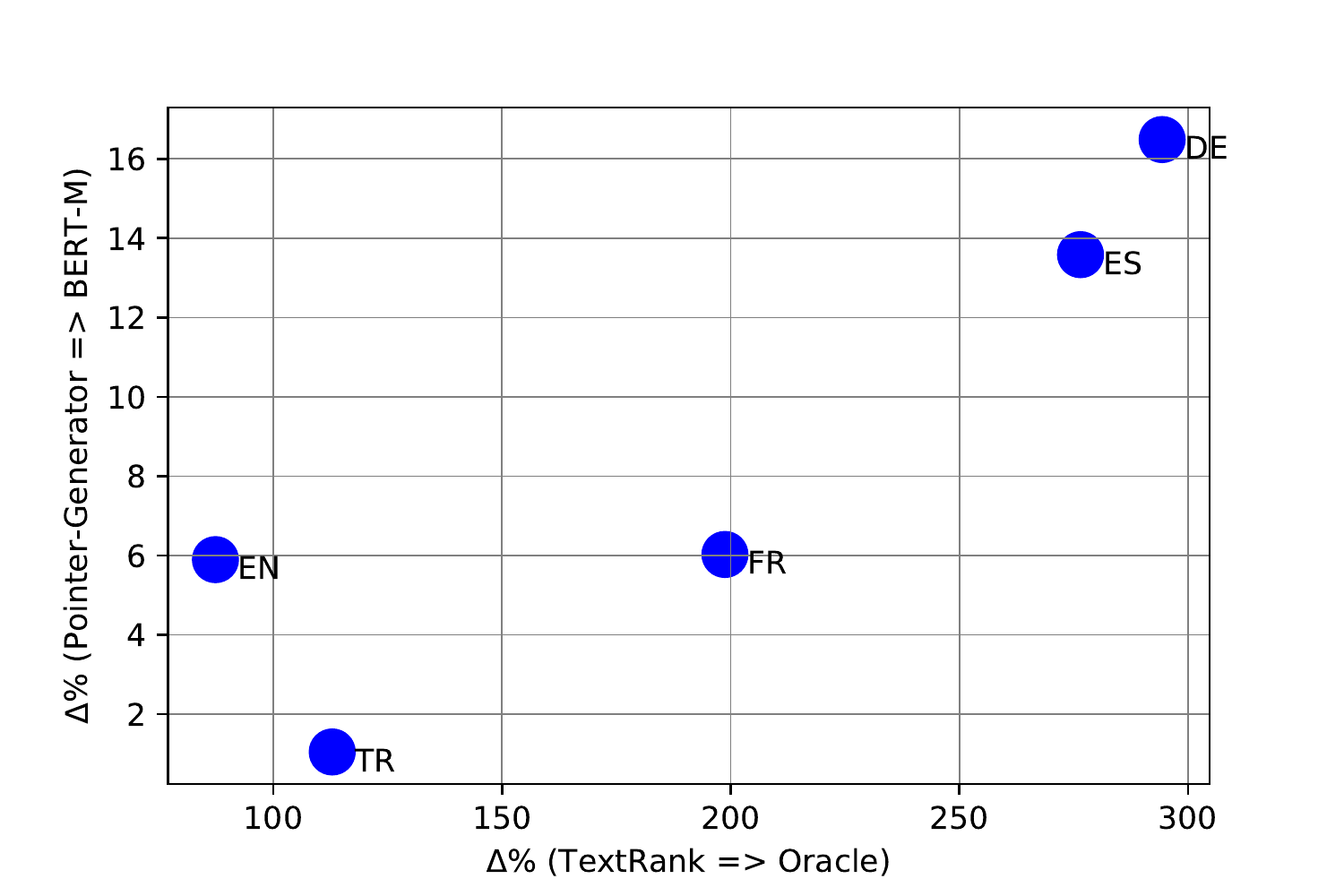}
    \caption{Improvement rates from TextRank to Oracle (in abscissa) against rates from  Pointer Generator to M-BERT (in ordinate).}
    \label{fig:delta_models}
\end{figure}

\paragraph{The benefits of pretraining}
We hypothesize that the closer an unsupervised model performance to its maximum limit, the less improvement would come from pretraining. In Figure~\ref{fig:delta_models}, we plot the improvement rate from TextRank to Oracle, against that of Pointer-Generator to M-BERT.

Looking at the correlation emerging from the plot, the hypothesis appears to hold true for all languages, including Russian -- not plotted for scaling reasons ($x=808;y=40$), with the exception of English. 
This exception is probably due to the aforementioned bias of TextRank towards the English language.

\paragraph{Pointer Generator and M-BERT}

Finally, we observe in our results that M-BERT always outperforms the Pointer Generator. However, the ratio is not homogeneous across the different languages as reported in Table~\ref{tab:ratios}. In particular, the improvement for  German is much more important than the one for French. Interestingly, this observation is in line with the results reported for Machine Translation: the Transformer \cite{vaswani2017attention} outperforms significantly ConvS2S \cite{gehring2017convolutional} for English to German but obtains comparable results for English to French -- see Table 2 in \citet{vaswani2017attention}. 

Neither model is pretrained, nor based on LSTM \cite{hochreiter1997long}, and they both use BPE tokenization \cite{shibata1999byte}. Therefore, the main difference is represented by the self-attention mechanism introduced in the Transformer, while ConvS2S used only source to target attention. We thus hypothesise that self-attention plays an important role for German but has a limited impact for French. This could find an explanation in the morphology of the two languages: 
in statistical parsing, \citet{tsarfaty2010statistical} considered German to be very sensitive to word order, due to its rich morphology, as opposed to French. Among other reasons, the flexibility of its syntactic ordering is mentioned. This corroborates the hypothesis that self-attention might help preserving information for languages with higher degrees of word order freedom.

\subsection{Possible derivative usages of MLSUM}

\paragraph{Multilingual Question Answering}
Originally, CNN/DM was a Question Answering dataset \cite{hermann2015teaching}. The hypothesis is that the information in the summary is also contained in the pair article. Hence, questions can be generated from the summary sentences by masking the Named Entities contained therein. 

The masked entities represent the answers, and thus a masked question should be answerable given the source article. So far, no multilingual \emph{training} dataset has been proposed for Question Answering. 

This methodology could be thus applied on MLSUM as a first step toward a large-scale multilingual Question Answering corpus. Incidentally, this would also allow progressing towards multilingual Question Generation, a crucial component to employ the neural summarization metrics mentioned in Section~\ref{sec:evaluation_metrics}.

\paragraph{News Title Generation}
While the release of MLSUM hereby described covers only article-summary pairs, the archived news articles also include the corresponding titles. The accompanying code for parsing the articles allows to easily retrieve the titles and thus use them for News Title Generation.

\paragraph{Topic detection}
A topic/category can be associated with each article/summary pair, by simply parsing the corresponding URL. 
A natural application of this data for summarization would be for template based summarization \cite{perez2019generating}, using it as additional features. However, it can also be a useful multilingual resource for topic detection.

\section{Conclusion}
We presented MLSUM, the first large-scale MultiLingual SUMmarization dataset, comprising over 1.5M article/summary pairs in French, German, Russian, Spanish, and Turkish.
We detailed its construction, and its complementary nature to the CNN/DM summarization dataset for English. We reported extensive preliminary experiments, highlighting biases observed in existing summarization models as well as analyzing and investigating the relative performances across languages of state-of-the-art approaches.
In future work, we plan to add other languages including Arabic and Hindi, and to investigate the adaptation of neural metrics to multilingual summarization.

\bibliography{acl2020}
\bibliographystyle{acl_natbib}
\clearpage
\appendix
\onecolumn
\section{Samples}
\label{sec:appendix}

\begin{center}
\noindent\fbox{%
    \parbox{.96\linewidth}{
        {\small 
        \begin{center}
        \textbf{-- FRENCH --}
        \end{center}
        
        \textbf{summary}
        Terre d'origine du clan Karzaï, la ville méridionale de Kandahar est aussi un bastion historique des talibans, où le mollah Omar a vécu et conservé de profondes racines. C'est sur cette terre pachtoune, plus qu'à Kaboul, que l'avenir à long terme du pays pourrait se décider.
        
        \textbf{body} 
        Lorsque l'on parle de l'Afghanistan, les yeux du monde sont rivés sur sa capitale, Kaboul. C'est là que se concentrent les lieux de pouvoir et où se détermine, en principe, son avenir. C'est aussi là que sont réunis les commandements des forces civiles et militaires internationales envoyées sur le sol afghan pour lutter contre l'insurrection et aider le pays à se reconstruire. Mais, à y regarder de plus près, Kaboul n'est qu'une façade. Face à un Etat inexistant, une structure du pouvoir afghan encore clanique, des tribus restées puissantes face à une démocratie artificielle importée de l'extérieur, la vraie légitimité ne vient pas de Kaboul. La géographie du pouvoir afghan aujourd'hui oblige à dire qu'une bonne partie des clés du destin de la population afghane se trouve au sud, en terre pachtoune, dans une cité hostile aux étrangers, foyer historique des talibans, Kandahar. Kandahar est la terre d'origine du clan Karzaï et de sa tribu, les Popalzaï. Hamid Karzaï, président afghan, tient son pouvoir du poids de son clan dans la région. Mi-novembre 2009, dans la grande maison de son frère, Wali, à Kandahar, se pressaient des chefs de tribu venus de tout l'Afghanistan, les piliers de son réseau. L'objet de la rencontre : faire le bilan post-électoral après la réélection contestée de son frère à la tête du pays. Parfois décrié pour ses liens supposés avec la CIA et des trafiquants de drogue, Wali Karzaï joue un rôle politique méconnu. Il a organisé la campagne de son frère, et ce jour-là, à Kandahar, se jouait, sous sa houlette, l'avenir de ceux qui avaient soutenu ou au contraire refusé leur soutien à Hamid. Chef d'orchestre chargé du clan du président, Wali est la personnalité forte du sud du pays. Les Karzaï adossent leur influence à celle de Kandahar dans l'histoire de l'Afghanistan. Lorsque Ahmad Shah, le fondateur du pays, en 1747, conquit la ville, il en fit sa capitale. "Jusqu'en 1979, lors de l'invasion soviétique, Kandahar a incarné le mythe de la création de l'Etat afghan, les Kandaharis considèrent qu'ils ont un droit divin à diriger le pays", résume Mariam Abou Zahab, experte du monde pachtoune. "Kandahar, c'est l'Afghanistan, explique à ceux qui l'interrogent Tooryalaï Wesa, gouverneur de la province. La politique s'y fait et, encore aujourd'hui, la politique sera dictée par les événements qui s'y dérouleront." Cette emprise de Kandahar s'évalue aux places prises au sein du gouvernement par "ceux du Sud". La composition du nouveau gouvernement, le 19 décembre, n'a pas changé la donne. D'autant moins que les rivaux des Karzaï, dans le Sud ou ailleurs, n'ont pas réussi à se renforcer au cours du dernier mandat du président. L'autre terre pachtoune, le grand Paktia, dans le sud-est du pays, à la frontière avec le Pakistan, qui a fourni tant de rois, ne dispose plus de ses relais dans la capitale. Kandahar pèse aussi sur l'avenir du pays, car s'y trouve le coeur de l'insurrection qui menace le pouvoir en place. L'OTAN, défiée depuis huit ans, n'a cessé de perdre du terrain dans le Sud, où les insurgés contrôlent des zones entières. Les provinces du Helmand et de Kandahar sont les zones les plus meurtrières pour la coalition et l'OTAN semble dépourvue de stratégie cohérente. Kandahar est la terre natale des talibans. Ils sont nés dans les campagnes du Helmand et de Kandahar, et le mouvement taliban s'est constitué dans la ville de Kandahar, où vivait leur chef spirituel, le mollah Omar, et où il a conservé de profondes racines. La pression sur la vie quotidienne des Afghans est croissante. Les talibans suppléent même le gouvernement dans des domaines tels que la justice quotidienne. Ceux qui collaborent avec les étrangers sont stigmatisés, menacés, voire tués. En guise de premier avertissement, les talibans collent, la nuit, des lettres sur les portes des "collabos". "La progression talibane est un fait dans le Sud, relate Alex Strick van Linschoten, unique spécialiste occidental de la région et du mouvement taliban à vivre à Kandahar sans protection. L'insécurité, l'absence de travail poussent vers Kaboul ceux qui ont un peu d'éducation et de compétence, seuls restent les pauvres et ceux qui veulent faire de l'argent." En réaction à cette détérioration, les Américains ont décidé, sans l'assumer ouvertement, de reprendre le contrôle de situations confiées officiellement par l'OTAN aux Britanniques dans le Helmand et aux Canadiens dans la province de Kandahar. Le mouvement a été progressif, mais, depuis un an, les Etats-Unis n'ont cessé d'envoyer des renforts américains, au point d'exercer aujourd'hui de fait la direction des opérations dans cette région. Une tendance qui se renforcera encore avec l'arrivée des troupes supplémentaires promises par Barack Obama. L'histoire a montré que, pour gagner en Afghanistan, il fallait tenir les campagnes de Kandahar. Les Britanniques l'ont expérimenté de façon cuisante lors de la seconde guerre anglo-afghane à la fin du XIXe siècle et les Soviétiques n'en sont jamais venus à bout. "On sait comment cela s'est terminé pour eux, on va essayer d'éviter de faire les mêmes erreurs", observait, mi-novembre, optimiste, un officier supérieur américain. 
        }
    }%
}

\noindent\fbox{%
    \parbox{.96\linewidth}{
        {\small 
        \begin{center}
        \textbf{-- GERMAN --}
        \end{center}
        
        \textbf{summary}
        Die Wurzeln des Elends liegen in der Vergangenheit. Haiti bezahlt immer noch für seine Befreiung vor 200 Jahren. Auch damals nahmen die Wichtigen der Welt den Insel-Staat nicht ernst.
        
        \textbf{body} 
Das Portrait von 1791 zeigt Haitis Nationalhelden François-Dominique Toussaint L'Ouverture. Er war einer der Anführer der Revolution in Haiti und Autor der ersten Verfassung. Die Wurzeln des Elends liegen in der Vergangenheit. Haiti bezahlt immer noch für seine Befreiung vor 200 Jahren. Auch damals nahmen die Wichtigen der Welt den Insel-Staat nicht ernst. Am vergangenen Wochenende schickte der britische Architekt und Gründer der Organisation Architecture for Humanity eine atemlose, verzweifelte E-Mail an seine Freunde und Unterstützer. "Nicht Erdbeben, sondern Gebäude töten Menschen" schrieb er in die Betreffzeile. Damit brachte er auf den Punkt, was auch der Geologe und Autor Simon Winchester oder der Urbanist Mike Davis immer wieder geschrieben haben - es gibt keine Naturkatastrophen. Es gibt nur gewaltige Naturereignisse, die tödliche Folgen haben. Die Konsequenz aus dieser Schlussfolgerung ist die Schuldfrage. Einfach lässt sie sich beantworten: Gier und Korruption sind fast immer die Auslöser einer Katastrophe. In Haiti aber liegen die Wurzeln der Tragödie tief in der Geschichte des Landes. Diese begann nach europäischer Rechnung im Jahre 1492, als Christopher Kolumbus auf der Insel landete, die ihre Ureinwohner Aytí nannten. Kolumbus benannte die Insel in Hispaniola um und gründete mit den Trümmern der gestrandeten Santa Maria die erste spanische Kolonie in der Neuen Welt. Ende des 17. Jahrhunderts besetzten französische Siedler den Westen der Insel, den Frankreich 1691 zur französischen Kolonie Sainte Domingue erklärte. Ideale der Französischen Revolution Gut hundert Jahre währte die Herrschaft der beiden Kolonialherren über die geteilte Insel. "Saint Domingue war die reichste europäische Kolonie in den Amerikas", schrieb der Historiker Hans Schmidt. 1789 kam fast die Hälfte des weltweit produzierten Zuckers aus der französischen Kolonie, die auch in der Produktion von Kaffee, Baumwolle und Indigo Weltmarktführer war. 450000 Sklaven arbeiteten auf den Plantagen, und sie erfuhren bald vom neuen Geist ihrer Herren. Die Französische Revolution brachte die Ideale von Freiheit, Gleichheit und Brüderlichkeit in die Karibik. Im August 1791 war es so weit. Der Voodoo-Priester Dutty Boukman rief während einer Messe zum Aufstand. Einer der erfolgreichsten Kommandeure der Rebellion war der ehemalige Sklave François-Dominique Toussaint L'Ouverture, nach dem heute der Flughafen von Port-au-Prince benannt ist. 1801 gab Toussaint dem Land seine erste Verfassung, die gleichzeitig eine Unabhängigkeitserklärung war. Für Napoleon sollte Haiti eine Schmach bleiben. Daraufhin sandte Napoleon Bonaparte Kriegsschiffe und Soldaten. Toussaint wurde verhaftet und nach Frankreich gebracht, wo er im Kerker starb. Doch als Napoleon im Jahr darauf die Sklaverei wieder einführen wollte, kam es erneut zum Aufstand. Verzweifelt baten die französischen Truppen im Sommer 1803 um Verstärkung. Da aber hatte Napoleon schon das Interesse an der Neuen Welt verloren. Im April hatte er seine Kolonie Louisiana an die Nordamerikaner verkauft, ein Gebiet, das rund ein Viertel des Staatsgebietes der heutigen USA umfasste. Für Napoleon sollte Haiti eine Schmach bleiben. Am 1. Januar 1804 erklärte der Rebellenführer Jean-Jacques Dessalines, die ehemalige Kolonie heiße nun Haiti und sei eine freie Republik. Der erste und bis zur Abschaffung der Sklaverei einzige erfolgreiche Sklavenaufstand der Neuen Welt war ein Schock für die Großmächte der Kolonialära, die ihren Reichtum auf der Sklaverei gegründet hatten. Ein Handel, der die Geschichte Haitis bis heute bestimmt Die Freiheit hatte ihren Preis. Ein Großteil der Plantagen war zerstört, ein Drittel der Bevölkerung Haitis den Kämpfen zum Opfer gefallen. Vor allem aber wollte keine Kolonialmacht die junge Republik anerkennen. Im Gegenteil -die meisten Länder unterstützten das Embargo der Insel und die Forderungen französischer Sklavenherren nach Reparationszahlungen. In der Hoffnung, als freie Nation Zugang zu den Weltmärkten zu erhalten, ließ sich die neue Machtelite Haitis auf einen Handel ein, der die Geschichte der Insel bis heute bestimmt. Mehr als zwei Jahrzehnte nach dem Sieg der Rebellen entsandte König Karl X. seine Kriegsschiffe nach Haiti. Ein Emissär stellte die Regierung vor die Wahl: Haiti sollte für die Anerkennung als Staat 150 Millionen Francs bezahlen. Sonst würde man einmarschieren und die Bevölkerung erneut versklaven. Haiti nahm Schulden auf und bezahlte. Bis zum Jahre 1947 lähmte die Schuldenlast die haitianische Wirtschaft und legte den Grundstein für Armut und Korruption. 2004 ließ der damalige haitianische Präsident Jean-Bertrand Aristide errechnen, was diese "Reparationszahlungen" für Haiti bedeuteten. Rund 22 Milliarden amerikanische Dollar Rückzahlung forderten seine Anwälte damals von der französischen Regierung. Vergebens. Lesen Sie auf der nächsten Seite, wie Haiti von den Akteuren der Weltbühne geschnitten wurde.
       }
    }%
}

\noindent\fbox{%
    \parbox{.96\linewidth}{
        {\small 
        \begin{center}
        \textbf{-- SPANISH --}
        \end{center}
        
        \textbf{summary}
        El aeropuerto ha estado hasta las 15.00 con sólo dos pistas por ausencia de 5 de los 18 controladores aéreos.- Varias aerolíneas han denunciado demoras de "hasta 60 minutos con los pasajeros embarcados"
        
        \textbf{body} 
        El espacio hará un repaso cronológico de la vida de la Esteban desde el momento en el que una completa desconocida comenzó a aparecer en los medios en 1998 como la novia de Jesulín de Ubrique hasta llegar a hoy en día, convertida en la princesa del pueblo, en concreto del popular madrileño distrito de San Blas donde vive, tal y como algunos la han calificado, y protagonista de portadas de revistas, diarios y portales web y de aparecer incluso entre los personajes más populares de Google. Junto a María Teresa Campos, estarán en el plató Patricia Pérez, presentadora del programa matinal de los sábados en Telecinco Vuélveme loca, quien ha conducido las campanadas en cuatro ocasiones, y los comentaristas Maribel Escalona, Emilio Pineda y José Manuel Parada.Los vuelos han venido registrando este viernes importantes retrasos en Barajas a pesar de que desde las 15.00 el aeropuerto opera con las cuatro pistas, según han informado fuentes de AENA, mientras las compañías han denunciado demoras por parte de los controladores de hasta 60 minutos con los pasajeros embarcados. Según los datos facilitados por AENA, la ausencia por la mañana de 5 de los 18 controladores que estaban programados en el turno de la torre de control de Barajas obligó a cerrar dos de las pistas del aeropuerto, lo que generó retrasos medios de 30 minutos.
        }
    }%
}

\vspace{10em}

\noindent\fbox{%
    \parbox{.96\linewidth}{
        {\small 
        \begin{center}
        \textbf{-- TURKISH --}
        \end{center}
        
        \textbf{summary}
        Ataması yapılmayan öğretmenler miting yaptı. Öğretmen adaylarına Muharrem İnce ve TEKEL işçileri de destek verdi.
        
        \textbf{body} 
        Yetersiz açılan kadrolar nedeniyle ataması yapılamayan öğretmen adayları Ankara'da miting yaptı. Tekel işçilerinin de destek verdiği öğretmen adaylarının mitinginde öğretmen kökenli CHP Milletvekili Muharrem İnce de hazır bulundu. Türkiye'nin çeşitli illerinden gelen ''Ataması Yapılmayan Öğretmenler Platformu'' üyesi sözleşmeli öğretmenler, öğle saatlerinde Abdi İpekçi Parkı'nda toplandı. ''Milletvekilliği için KPSS getirilsin'', ''1 kadrolu öğretmen = 3 ücretli öğretmen'' ve ''Ücretli köle olmayacağız'' yazılı dövizler taşıyan ve aynı içerikli sloganlar atan öğretmenlerin düzenlediği mitinge, bazı siyasi parti, sivil toplum kuruluşu temsilcileri ve TEKEL işçileri de destek verdi. CHP Yalova Milletvekili Muharrem İnce, okullarda derslerin boş geçtiğini öne sürerek, ''Okullar öğretmensiz, öğretmenler ise işsiz'' dedi. Hükümetin bu gençlerin sesini duyması gerektiğini belirten İnce, ''Bu ülkenin 250 bin eğitim fakültesi mezunu genci iş bekliyorsa bu hükümetin ve ülkenin ayıbıdır. Eğitim sorununu çözememiş bir hükümet bu ülkenin hiçbir sorununu çözememiş demektir. Bu kadar önemli bir soruna kulaklarını tıkayamaz'' diye konuştu. ''Ankara'nın göbeğinde derslerin boş geçtiğini'' ileri süren İnce, ''Bu ülkede fizik ve matematik öğretmeni atanmıyor ama bunların 100 katı din dersi öğretmeni atanıyor'' dedi. Platform adına yapılan açıklamada da Türkiye'de her yıl üniversite bitirerek diplomasını alan öğretmenlerin eğitim alanındaki yetersizlik dolayısıyla işsizler kervanına katıldığı ifade edildi. Talep edilen hakların insancıl ve makul olduğu belirtilen açıklamada, öğretmenlerin haklarını vermeyenlerin kötü niyetli olduğu öne sürüldü. Açıklamada, hükümetin eğitim politikası eleştirilerek, sözleşmeli öğretmenlerin kadrolu atamalarının yapılması, öğretmen yetiştiren fakültelere öğretmen ihtiyacı kadar öğretmen adayı alınması ve KPSS yerine daha şeffaf bir atama sistemi getirilmesi istendi. ÖLÜM ORUCU BAŞLATACAKLAR Çeşitli sivil toplum kuruluşu temsilcilerinin de konuştuğu mitingde, kadrolu atamalar yapılmadığı takdirde iş bırakma eylemi ve ölüm orucu yapılacağı duyuruldu.
        }
    }%
}

\clearpage

\end{center}


\end{document}